\documentclass[sigconf]{acmart}

\AtBeginDocument{%
  \providecommand\BibTeX{{%
    \normalfont B\kern-0.5em{\scshape i\kern-0.25em b}\kern-0.8em\TeX}}}

\copyrightyear{2022} 
\acmYear{2022} 
\setcopyright{acmcopyright}\acmConference[WSDM '22]{Proceedings of the Fifteenth ACM International Conference on Web Search and Data Mining}{February 21--25, 2022}{Tempe, AZ, USA}
\acmBooktitle{Proceedings of the Fifteenth ACM International Conference on Web Search and Data Mining (WSDM '22), February 21--25, 2022, Tempe, AZ, USA}
\acmPrice{15.00}
\acmDOI{10.1145/3488560.3502183}
\acmISBN{978-1-4503-9132-0/22/02}

\usepackage{multirow}
\usepackage{enumitem}
\usepackage{arydshln}

\begin{document}
\fancyhead{}

\title{Node Co-occurrence based Graph Neural Networks for Knowledge Graph Link Prediction}

\author{Dai Quoc Nguyen}
\affiliation{\institution{Oracle Labs, Australia}}
\email{dai.nguyen@oracle.com}
\authornote{The first two authors contributed equally to this work. \\ This work was done before Dai Quoc Nguyen joined Oracle Labs, Australia.}

\author{Vinh Tong}
\affiliation{\institution{VinAI Research, Vietnam}}
\email{v.vinhtv4@vinai.io}
\authornotemark[1]

\author{Dinh Phung}
\affiliation{\institution{Monash University, Australia}}
\email{dinh.phung@monash.edu}

\author{Dat Quoc Nguyen}
\affiliation{\institution{VinAI Research, Vietnam}}
\email{v.datnq9@vinai.io}

\begin{abstract}

We introduce a novel embedding model, named NoGE, which aims to integrate co-occurrence among entities and relations into graph neural networks to improve knowledge graph completion (i.e., link prediction). Given a knowledge graph, NoGE constructs a single graph considering entities and relations as individual nodes. NoGE then computes weights for edges among nodes based on the co-occurrence of entities and relations. Next, NoGE proposes Dual Quaternion Graph Neural Networks (DualQGNN) and utilizes DualQGNN to update vector representations for entity and relation nodes. NoGE then adopts a score function to produce the triple scores. Comprehensive experimental results show that NoGE obtains state-of-the-art results on three new and difficult benchmark datasets CoDEx for knowledge graph completion. 


\end{abstract}

\ccsdesc[500]{Computing methodologies~Natural language processing}
\ccsdesc[500]{Computing methodologies~Neural networks}

\keywords{graph neural networks, knowledge graph completion, quaternion}

\maketitle

\section{Introduction}

Knowledge graphs (KGs)---representing relationships among entities in the form of triples \textit{(head, relation, tail)} denoted as \textit{(h, r, t)}---are useful resources for many NLP and information retrieval applications such as semantic search and question answering \citep{8047276}. 
However, large knowledge graphs
are still incomplete 
\citep{bordes2011learning,West:2014}. 
Therefore, many research works have focused on inferring missing triples in KGs, i.e., predicting whether a triple not in KGs is likely to be valid or not \citep{Lao2010,Nguyen2020KGC,NGUYEN2021Thesis}. Consequently, many embedding models have been proposed to learn vector representations for entities and relations and return a score for each triple, such that valid triples have higher scores than invalid ones \citep{NIPS2013_5071,NIPS2013_5028}. 
For example, the score of the valid triple (Melbourne, city\_of, Australia) is higher than the score of the invalid one (Melbourne, city\_of, Vietnam).

In addition to conventional KG embedding models such as TransE \citep{NIPS2013_5071}, DistMult \citep{Yang2015},  ComplEx \citep{Trouillon2016}, ConvE \cite{Dettmers2017}, ConvKB \citep{Nguyen2018,Nguyen2018ConvKBfull}, and TuckER \cite{balazevic2019tucker}, recent approaches have adapted graph neural networks (GNNs) for knowledge graph completion \citep{schlichtkrull2017modeling,shang2019end,vashishth2020compositionbased,Nguyen2020QGNN}. 
In general, vanilla GNNs are modified and utilized as an encoder module to update vector representations for entities and relations; then these vector representations are fed into a decoder module that adopts a score function (e.g., as employed in TransE, DistMult, and ConvE) to return the triple scores. 
Those GNN-based models, however, are still outperformed by other conventional models  on some benchmark datasets \cite{Nguyen2020KGC}.
To boost the model performance, our motivation comes from the fact that {entities and relations forming facts often co-occur frequently in news articles, texts, and documents, e.g., ``Melbourne'' co-occurs frequently together with ``Australia''.}

We thus propose a new effective GNN-based KG embedding model, named NoGE, to integrate co-occurrence among entities and relations in the encoder module for knowledge graph completion (as the \textit{first contribution}).
NoGE is different from other existing GNN-based KG embedding models in two important aspects: 
(i) Given a knowledge graph, NoGE builds a single graph, which contains entities and relations as individual nodes; (ii) NoGE counts the co-occurrence of entities and relations to compute weights of edges among nodes, resulting in a new weighted adjacency matrix.
Consequently, NoGE can  leverage the vanilla GNNs directly on the single graph of entity and relation nodes associated with the new weighted adjacency matrix.
As the \textit{second contribution}, NoGE also proposes a novel form of GNNs, named Dual Quaternion Graph Neural Networks (DualQGNN) as the encoder module. 
Then NoGE  employs a score function, e.g.  QuatE \citep{zhang2019quaternion}, as the decoder module to return the triple scores. As our \textit{final contribution}, we conduct extensive experiments to compare our NoGE with other strong GNN-based baselines and show that  NoGE outperforms these baselines as well as other up-to-date KG embedding models and obtains state-of-the-art results on three new and difficult benchmark datasets CoDEx-S, CoDEx-M, and CoDEx-L \citep{safavi2020codex} for knowledge graph completion.

\section{Background}

\subsection{Related work}

We represent each single graph $\mathcal{G} = \left(\mathcal{V}, \mathit{E}\right)$, where $\mathcal{V}$ is a set of nodes and $\mathit{E}$ is a set of edges.
Graph Convolutional Networks (GCNs) \citep{kipf2017semi} update vector representations for nodes $\mathsf{v} \in \mathcal{V}$ via using multiple layers stacked on top of each other.
Regarding the GNN-based KG embedding approaches, R-GCN \citep{schlichtkrull2017modeling} modifies GCNs to introduce a specific encoder to update only entity embeddings.
R-GCN then uses DistMult as its decoder module.
Recently, CompGCN \citep{vashishth2020compositionbased} customizes GCNs to consider composition operations 
between entities and relations in the encoder module.
CompGCN then applies ConvE \citep{Dettmers2017} as the decoder module.
Note that R-GCN and CompGCN do not consider co-occurrence among entities and relations in the encoder module. This limitation also exists in other GNN-based models such as SACN \citep{shang2019end}.
Therefore, arguably this could lower the performance of these existing GNN-based models.
One of our  key contributions is to integrate co-occurrence among entities and relations in the encoder module.

\subsection{Dual quaternion background}

A background in quaternion can be found in recent works \citep{zhang2019quaternion,Nguyen2020QGNN}. We briefly provide a background in dual quaternion \citep{clifford1871}. 
A dual quaternion $h \in \mathbb{H}_d$ is given in the form: 
$h = q + \epsilon p$,
where $q$ and $p$ are quaternions $\in \mathbb{H}$, $\epsilon$ is the dual unit with $\epsilon^2 = 0$. 

\paragraph{Conjugate.} The conjugate $h^\ast$ of a dual quaternion $h$ is defined as:
$h^\ast = q^\ast + \epsilon p^\ast$.

\paragraph{Addition.} 
The addition of two dual quaternions $h_1 = q_1 + \epsilon p_1$ and $h_2 = q_2 + \epsilon p_2$ is defined as:
$h_1 + h_2 = (q_1 + q_2) + \epsilon(p_1 + p_2)$.

\paragraph{Dual quaternion multiplication.} The dual quaternion multiplication $\otimes_d$ of two dual quaternions $h_1$ and $h_2$ is defined as:
\begin{equation}
h_1 \otimes_d h_2 = (q_1\otimes q_2) + \epsilon(q_1\otimes p_2 + p_1 \otimes q_2) \nonumber
\end{equation}
where $\otimes$ denotes the Hamilton product between two quaternions.

\paragraph{Norm.} The norm $\|h\|$ of a dual quaternion $h$ is a dual number, which is usually defined as:
$\|h\| = \sqrt{h\otimes_dh^\ast} = \sqrt{\|q\|^2 + 2\epsilon q\bullet p}
= \|q\| + \epsilon\frac{q\bullet p}{\|q\|}$.

\paragraph{Unit dual quaternion.} A dual quaternion $h$ is \textit{unit} if $h\otimes_dh^\ast = 1$ with $\|q\|^2 = 1$ and $q\bullet p = 0$.

\paragraph{Normalization.} The normalized dual quaternion $h^\triangleleft$ is usually defined as:
$h^\triangleleft = \frac{h}{\|h\|} = q^\triangleleft + \epsilon\left(\frac{p}{\|q\|} - q^\triangleleft\frac{q\bullet p}{\|q\|^2}\right)$.

\paragraph{Matrix-vector multiplication.} The dual quaternion multiplication $\otimes_d$ of a dual quaternion matrix $\textbf{W}^{DQ} = \textbf{W}^Q_q + \epsilon \textbf{W}^Q_p$ and a dual quaternion vector $\boldsymbol{h}^{DQ} = \boldsymbol{q}^Q + \epsilon \boldsymbol{p}^Q$ is defined as:
\begin{equation} 
\textbf{W}^{DQ} \otimes_d \boldsymbol{h}^{DQ} =
(\textbf{W}^Q_q\otimes \boldsymbol{q}^Q) + \epsilon(\textbf{W}^Q_q\otimes \boldsymbol{p}^Q + \textbf{W}^Q_p \otimes \boldsymbol{q}^Q) \nonumber
\end{equation}
where the superscripts $^{DQ}$ and $^Q$ denote the dual Quaternion space $\mathbb{H}_d$ and the Quaternion space $\mathbb{H}$, respectively.

\section{Our proposed NoGE}

\begin{figure}[!t]
	\centering
	\includegraphics[width=0.45\textwidth]{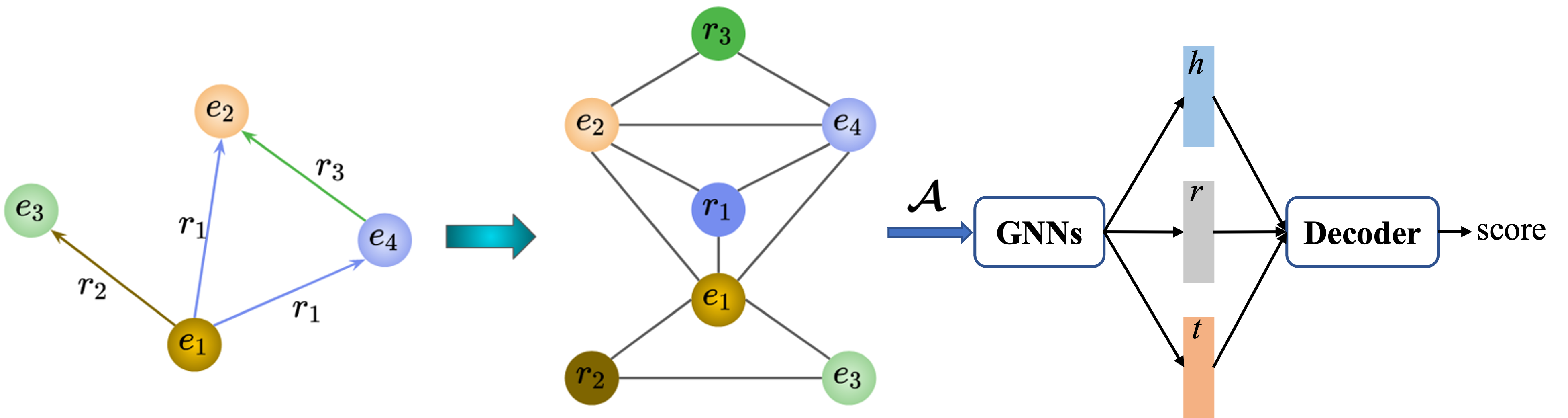}
    \caption{An illustration of our proposed NoGE.}
    \label{fig:graph_complex}
 \end{figure}
 
A knowledge graph $\mathit{G}$ is a collection of valid factual triples in the form of \textit{(head, relation, tail)} denoted as $(h, r, t)$ with $h, t \in \mathcal{E}$ and $r \in \mathcal{R}$, wherein $\mathcal{E}$ is a set of entities and $\mathcal{R}$ is a set of relations.
KG embedding models aim to embed entities and relations to a low-dimensional vector space and define a score function to give a score for each triple, such that the valid triples obtain higher scores than the invalid triples.

To enhance the efficiency of the encoder module, 
our motivation comes from the fact that entities and relations forming facts often co-occur  frequently in news articles, texts, and documents, e.g., ``Melbourne'' co-occurs together with ``city\_of'' frequently.
Given a knowledge graph $\mathit{G}$,  
NoGE builds a single graph $\mathcal{G}$ that contains entities and relations as nodes following Levi graph transformation \citep{levi1942finite}, as illustrated in Figure \ref{fig:graph_complex}. 
The total number of nodes in $\mathcal{G}$ is the sum of the numbers of entities and relations, i.e. $|\mathcal{V}|$ = $|\mathcal{E}|$ + $|\mathcal{R}|$. 
NoGE then builds edges among nodes based on the co-occurrence of entities and relations within the triples in $\mathit{G}$.
Formally, NoGE computes the weights of edges among nodes $\mathsf{v}$ and  $\mathsf{u}$ to create a new weighted adjacency matrix $\boldsymbol{\mathcal{A}}$ as follows:
\begin{equation}
\mathcal{A}_{\mathsf{v},\mathsf{u}} = \begin{cases}
\frac{p(\mathsf{v}, \mathsf{u})}{p(\mathsf{v})} & \text{if $\mathsf{v}$ and $\mathsf{u}$ are entity nodes, and}\\
& \text{\ \ \ \ \ $p(\mathsf{v}$, $\mathsf{u})$ $>$ 0}\\
p(\mathsf{v}, \mathsf{u}) & \text{if either of $\mathsf{v}$ and $\mathsf{u}$ is a relation}\\
& \text{\ \ \ \ \  node, and $p(\mathsf{v}$, $\mathsf{u})$ $>$ 0}\\
1 & \text{if $\mathsf{v}$ = $\mathsf{u}$} \\
0 & \text{otherwise}
\end{cases} \nonumber
\label{equa:newadjmatrix}
\end{equation}
wherein $p(\mathsf{v},\mathsf{u})$ and $p(\mathsf{v})$ are computed as:
\begin{equation}
p(\mathsf{v}, \mathsf{u}) = \frac{\#C(\mathsf{v},\mathsf{u})}{\#C} \ \ ; \ \ \ \ p(\mathsf{v}) = \frac{\#C(\mathsf{v})}{\#C} \nonumber
\end{equation}
where $\#C(\mathsf{v},\mathsf{u})$ is the number of co-occurrence of two nodes $\mathsf{v}$ and $\mathsf{u}$ within the triples in $\mathit{G}$; $\#C(\mathsf{v})$ is the number of triples in $\mathit{G}$, that contain $\mathsf{v}$; and $\#C$ is the total number of triples in $\mathit{G}$ (i.e., $|\mathit{G}|$).
As a consequence, NoGE can leverage the vanilla GNNs \citep{kipf2017semi,Nguyen2019UGT,Nguyen2020QGNN}, directly on  $\mathcal{G}$ and our newly proposed $\boldsymbol{\mathcal{A}}$.

Compared to the quaternion space \citep{hamilton1844ii}, the dual quaternion space \citep{clifford1871} has several advantages in  modeling rotations and translations, and efficiently representing rigid transformations \citep{torsello2011multiview}.
Therefore, we introduce Dual Quaternion Graph Neural Networks (DualQGNN) and then utilize our DualQGNN as the encoder module in NoGE as:
\begin{equation}
\boldsymbol{v}_{\mathsf{v}}^{(k+1),DQ} = \mathsf{g}\left(\sum_{\mathsf{u} \in \mathcal{N}_\mathsf{v}\cup\left\{\mathsf{v}\right\}}\mathsf{a}_{\mathsf{v},\mathsf{u}}\textbf{W}^{(k),DQ}\otimes_d\boldsymbol{v}_{\mathsf{u}}^{(k),DQ}\right) \nonumber
\label{equa:QGNN}
\end{equation}
where the superscript $^{DQ}$ denotes the dual Quaternion space $\mathbb{H}_d$; $\textbf{W}^{(k),DQ}$ is a dual quaternion weight matrix;  
$\otimes_d$ denotes the dual quaternion multiplication; and $\mathsf{g}$ can be a nonlinear activation function such as $\mathsf{tanh}$; $\boldsymbol{v}_\mathsf{u}^{(0),DQ} \in \mathbb{H}_d^n$ is an input vector for node $\mathsf{u}$, which is initialized and learned during training.
Importantly, $\mathsf{a}_{\mathsf{v},\mathsf{u}}$ is now an edge constant between nodes $\mathsf{v}$ and $\mathsf{u}$ in the re-normalized adjacency matrix $\tilde{\textbf{D}}^{-\frac{1}{2}}\widetilde{\boldsymbol{\mathcal{A}}}\tilde{\textbf{D}}^{-\frac{1}{2}}$, wherein $\widetilde{\boldsymbol{\mathcal{A}}} = \boldsymbol{\mathcal{A}} + \textbf{I}$, and $\tilde{\textbf{D}}$ is the diagonal node degree matrix of $\widetilde{\boldsymbol{\mathcal{A}}}$.

NoGE obtains the dual quaternion vector representations of entities and relations from the last DualQGNN layer of the encoder module. 
For each obtained dual quaternion representation, NoGE concatenates its two quaternion coefficients to produce a final quaternion representation.
These final quaternion representations of entities and relations are then fed to QuatE \citep{zhang2019quaternion}, employed as the decoder module, to compute the score of \textit{(h, r, t)} as:
\begin{equation}
f(h, r, t) = \left(\boldsymbol{v}_h^Q \otimes \boldsymbol{v}_r^{\triangleleft,Q}\right) \bullet \boldsymbol{v}_t^Q \nonumber
\label{equa:quate}
\end{equation}
where the superscript $^Q$ denotes the Quaternion space $\mathbb{H}$; $\otimes$ denotes the Hamilton product; $^\triangleleft$ denotes the normalized quaternion; and $\bullet$ denotes the quaternion-inner product.

We then apply the Adam optimizer \citep{kingma2014adam} to train our proposed NoGE by minimizing the binary cross-entropy loss function \citep{Dettmers2017}  as:
{\small
\begin{equation}
\mathcal{L} = -\sum_{\substack{(h, r, t) \in \{\mathit{G} \cup \mathit{G}'\}}}\left(l_{(h, r, t)}\log\left(p_{(h, r, t)}\right)\right. + 
\left.\left(1 - l_{(h, r, t)}\right)\log\left(1 - p_{(h, r, t)}\right)\right) \nonumber
\label{equal:losscrossentropy}
\end{equation}
}
\begin{equation*}
\text{in which, } l_{(h, r, t)} = \left\{ 
  \begin{array}{l}
1 \ \ \ \text{for } (h, r, t)\in\mathit{G}\\
0 \ \ \ \text{for } (h, r, t)\in\mathit{G}'
  \end{array} \right.
\end{equation*}
where $p_{(h, r, t)} = \mathsf{sigmoid}\left(f(h, r, t)\right)$. $\mathit{G}$ and $\mathit{G}'$ are collections of valid and invalid triples, respectively.

\section{Experimental setup and results}
\label{ssec:kbc}

\subsection{Experimental setup} 
We evaluate our proposed NoGE for the knowledge graph completion task, i.e., link prediction \citep{NIPS2013_5071}, which aims to predict a missing entity given a relation with another entity, e.g., predicting a head entity $h$ given $(?, r, t)$ or predicting a tail entity $t$ given $(h, r, ?)$. 
The results are calculated by ranking the scores produced by the score function $f$ on triples in the test set.

\subsubsection{Datasets}
\citet{safavi2020codex} point out issues with existing KG completion datasets and thus introduce three new and more appropriately difficult benchmark datasets CoDEx-S, CoDEx-M, and CoDEx-L.
These three open-domain CoDEx datasets are derived from Wikidata and Wikipedia to cover more diverse and interpretable content and make a more challenging prediction task. Therefore, we employ these new datasets in our experiments.


\subsubsection{Evaluation protocol}
Following \citet{NIPS2013_5071}, for each valid test triple $(h, r, t)$, we replace either $h$ or $t$ by each of all other entities to create a set of corrupted triples.
We also use the ``{Filtered}'' setting protocol \citep{NIPS2013_5071}.
We rank the valid test triple and corrupted triples in descending order of their scores and report mean reciprocal rank (MRR) and Hits@$10$ (the proportion of the valid triples ranking in top $10$ predictions).
The final scores on the test set are reported for the model that obtains the highest MRR on the validation set. 


\subsubsection{Training protocol}

We set the same dimension value for both the embedding size and the hidden size of the DualQGNN hidden layers, wherein we vary the dimension value in \{32, 64, 128\}.
We fix the batch size to 1024.
We employ $\mathsf{tanh}$ for the nonlinear activation function $\mathsf{g}$.
We use the Adam optimizer \citep{kingma2014adam} to train our NoGE model up to 3,000 epochs on CoDEx-S and CoDEx-M, and 1,500 epochs on CoDEx-L. We use a grid search to choose the number of hidden layers  $\in \{1, 2, 3\}$ and the Adam initial learning rate  $\in \left\{1e^{-4}, 5e^{-4}, 1e^{-3}, 5e^{-3}\right\}$. To select the best checkpoint, we evaluate the MRR after each training epoch on the 
validation set.

\paragraph{Baselines' training protocol}


For other baseline models, we apply the same evaluation protocol. The training protocol is the same w.r.t. the optimizer, the hidden layers, the initial learning rate values, and the number of training epochs. 
In addition, we use the model-specific configuration for each baseline as follows:

\begin{itemize}[leftmargin=*]

    \item \textbf{QuatE} \citep{zhang2019quaternion}: We set the batch size to 1024 and vary the embedding dimension in \{64, 128, 256, 512\}. 

    \item Regarding the GNN-based baselines -- \textbf{R-GCN} \citep{schlichtkrull2017modeling}, \textbf{CompGCN} \citep{vashishth2020compositionbased}, \textbf{SACN} \citep{shang2019end}, and \textbf{our NoGE variants with QGNN and GCN} -- we also set the same dimension value for both the embedding size and the hidden size, wherein we vary the dimension value in \{64, 128, 256, 512\}. 
    
    \item \textbf{Our NoGE variant with QGNN}: This is a variant of our proposed method that utilizes QGNN \citep{Nguyen2020QGNN} as the encoder module.
    
    \item \textbf{Our NoGE variant with GCN}: This is a variant of our proposed method that utilizes GCN \citep{kipf2017semi} as the encoder module.
    
    \item \textbf{CompGCN}: We consider a CompGCN variant that set ConvE \cite{Dettmers2017} as its decoder module, circular-correlation as its composition operator, the kernel size to 7, and the number of output channels to 200, producing the best results as reported in the original implementation.  
    
    \item \textbf{SACN}: 
    For its decoder Conv-TransE, we set the kernel size to 5 and the number of output channels 200 as used in the original implementation. 

\end{itemize}

\subsection{Main results}

In Table \ref{tab:resultsCoDEx}, we report our obtained results  for NoGE and other strong baselines including QuatE \citep{zhang2019quaternion}, R-GCN \citep{schlichtkrull2017modeling}, SACN \citep{shang2019end} and CompGCN \citep{vashishth2020compositionbased} on the CoDEx datasets.

\begin{table}[!ht]
\centering
\caption{Experimental results on the CoDEx \textit{test} sets. Hits@10 (H@10) is reported in \%. 
The best scores are in {bold}, while the second best scores are in {underline}.
The results of TransE, ComplEx, ConvE, and TuckER are taken from \citep{safavi2020codex}.
The results of concurrent models SimRGCN and SimQGNN are taken from \citep{Nguyen2020QGNN}.
We get an out-of-memory for SACN on the large dataset CoDEx-L.
}
\resizebox{7.75cm}{!}{
\begin{tabular}{l|cc|cc|cc}
\hline
\multirow{2}{*}{\bf Method} & \multicolumn{2}{c|}{\bf CoDEx-S} & \multicolumn{2}{c|}{\bf CoDEx-M} & \multicolumn{2}{c}{\bf CoDEx-L} \\
\cline{2-7} 
& MRR   & H@10  & MRR   & H@10  & MRR   & H@10 \\
\hline
TransE  & 0.354 & 63.4 & 0.303 & 45.4 & 0.187 & 31.7 \\
ComplEx & \bf 0.465 & {64.6} & \underline{0.337} & {47.6} & 0.294 & 40.0 \\
ConvE   & {0.444} & 63.5 & 0.318 & 46.4 & 0.303 & 42.0 \\
TuckER  & {0.444} & 63.8 & {0.328} & 45.8 & {0.309} & {43.0} \\
\hdashline
SimRGCN & 0.427 & {64.7} & 0.322 & 47.5 & 0.307 & {43.2}\\
SimQGNN  & 0.435 & \textbf{65.2} & 0.323 & {47.7} & {0.310} & {43.7} \\
\hline
QuatE   & 0.449 & 64.4 & 0.323 & \underline{48.0} & \underline{0.312} & \underline{44.3}  \\
R-GCN   & 0.275 & 53.3 & 0.124 & 24.1 & 0.073 & 14.2 \\
SACN    & 0.374 & 59.4 & 0.294 & 44.3 & -- & --  \\
CompGCN & 0.395 & 62.1 & 0.312 & 45.7 & 0.304 & 42.8 \\
\hline
\textbf{NoGE}  & \underline{0.453} & \underline{65.0} & \bf 0.338 & \bf 48.4 & \bf 0.321 & \bf 45.0 \\
\hline
\end{tabular}
}
\label{tab:resultsCoDEx}
\end{table}

R-GCN is outperformed by all other models on these difficult benchmark datasets. 
This is similar to the findings mentioned in \citep{Dettmers2017,vashishth2020compositionbased}.
A possible reason is that R-GCN returns similar embeddings for different entities on the difficult benchmarks.
The recent model CompGCN uses ConvE as the decoder module, but it is outperformed by ConvE on CoDEx-S and CoDEx-M.
CompGCN also does not perform better than ComplEx and TuckER on the CoDEx datasets.
Similarly, QuatE, utilized as our NoGE's decoder module, also produces lower results than ComplEx, ConvE, and TuckER.

When comparing with QuatE and three other GNN-based baselines, our NoGE achieves substantial improvements on the CoDEx datasets.
For example, NoGE gains absolute Hits@10 improvements of 
2.9\%, 2.7\%, and 2.2\% over CompGCN on CoDEx-S, CoDEx-M, and CoDEx-L.
In general, our NoGE outperforms up-to-date embedding models and is considered as the best model on the CoDEx datasets. 
In particular, NoGE yields new state-of-the-art Hits@10 and MRR scores on CoDEx-M and CoDEx-L.

\paragraph{Ablation analysis} 
We compute and report our ablation results for three variants of NoGE in Table \ref{tab:ablationstudy}.
In general, the results degrade when using either QGNN or GCN as the encoder module, showing the advantage of our proposed DualQGNN. 
The scores also degrade when not using the new weighted adjacency matrix $\boldsymbol{\mathcal{A}}$. 
Besides, our NoGE variants with QGNN and GCN also substantially outperform three other GNN-based baselines R-GCN, SACN, and CompGCN, thus clearly showing the effectiveness of integrating our matrix $\boldsymbol{\mathcal{A}}$ into GNNs for KG completion.

\begin{table}[!ht]
\centering
\caption{Ablation results on the \textit{validation} sets. (i) NoGE variant utilizes QGNN as the encoder module instead of utilizing  our proposed encoder DualQGNN. (ii) NoGE variant utilizes GCN as the encoder module. 
(iii) NoGE variant re-uses the adjacency matrix $\textbf{A}$ rather than our proposed matrix $\boldsymbol{\mathcal{A}}$.}
\resizebox{7.75cm}{!}{
\begin{tabular}{l|cc|cc|cc}
\hline
\multirow{2}{*}{\bf Method} & \multicolumn{2}{c|}{\bf CoDEx-S} & \multicolumn{2}{c|}{\bf CoDEx-M} & \multicolumn{2}{c}{\bf CoDEx-L} \\
\cline{2-7}     & MRR   & H@10  & MRR   & H@10  & MRR   & H@10 \\
\hline
R-GCN & 0.287 & 54.7 & 0.122 & 23.8 & 0.073 & 14.1\\
SACN & 0.377 & 62.3 & 0.294 & 44.0 & -- & -- \\
CompGCN & 0.400 & 62.9 & 0.305 & 45.3 & 0.303 & 42.6\\
\hline
\textbf{NoGE}  & \bf 0.470 & \underline{65.6} & \bf 0.337 & \bf 48.1 & \bf 0.320 & \bf 44.6 \\
\hdashline
\ \ \ \ (i) w/ QGNN & \underline{0.465} & \bf 66.1 & \underline{0.332} & \underline{47.8} & \bf 0.320 & \underline{44.0} \\
\ \ \ \ (ii) w/ GCN & 0.445 & 65.3 & 0.325 & 47.3 & \underline{0.317} & 43.9 \\
\ \ \ \ (iii) w/o $\boldsymbol{\mathcal{A}}$ & 0.452 & 63.7 & 0.320 & 46.1 & 0.288 & 41.5\\
\hline
\end{tabular}
}
\label{tab:ablationstudy}
\end{table}

\section{Conclusion}
\label{sec:conclusion}

We have presented a novel model NoGE to integrate co-occurrence among entities and relations into graph neural networks for knowledge graph completion (i.e., link prediction).
Given a knowledge graph, NoGE constructs a single graph, which considers entities and relations as individual nodes.
NoGE builds edges among nodes based on the co-occurrence of entities and relations to create a new weighted adjacency matrix for the single graph, which can be  fed to vanilla GNNs.
NoGE then proposes new Dual Quaternion GNNs and utilizes a score function to obtain the triple scores.
NoGE obtains state-of-the-art performances on three new and difficult benchmark datasets CoDEx-S, CoDEx-M, and CoDEx-L for the knowledge graph completion task.
Our framework is available at: \url{https://github.com/daiquocnguyen/GNN-NoGE}, where we demonstrate the usage of different GNN encoders' implementations including GCN, QGNN and our DualQGNN as well as the usage of DistMult and QuatE as the decoder module.


\bibliographystyle{ACM-Reference-Format}
\bibliography{references}

\end{document}